\documentclass[conference,a4paper]{IEEEtran}

\usepackage{enumitem}
\usepackage{hyperref}
\usepackage{graphicx}
\usepackage{amsmath,amssymb,amsfonts,amsthm}
\usepackage[nameinlink,capitalise]{cleveref}
\usepackage{booktabs}
\usepackage[separate-uncertainty]{siunitx}
\usepackage[table]{xcolor}
\usepackage{tikz}
\usepackage{multirow}

\crefname{equation}{}{} 
\crefname{figure}{Fig.}{Fig.}
\Crefname{figure}{Fig.}{Fig.}

\theoremstyle{definition}
\newtheorem{definition}{Definition}
\crefname{definition}{Definition}{Definitions}
\newtheorem{property}{Property}
\crefname{property}{Property}{Properties}
\newenvironment{property*}[1]
  {\renewcommand\theproperty{#1}\property}
  {\endproperty}

\theoremstyle{plain}
\newtheorem{lemma}{Lemma}
\crefname{lemma}{Lemma}{Lemmas}
\newenvironment{lemma*}[1]
  {\renewcommand\thelemma{#1}\lemma}
  {\endlemma}

\crefname{proposition}{Proposition}{Propositions}

\def\BibTeX{{\rm B\kern-.05em{\sc i\kern-.025em b}\kern-.08em
    T\kern-.1667em\lower.7ex\hbox{E}\kern-.125emX}}

\begin{document}

\newcommand{\numsml}[1]{\small{\num[retain-zero-exponent,retain-explicit-plus]{#1}}}
\newcommand{\numbf}[1]{\fontseries{b}\selectfont{\num[detect-weight,mode=text]{#1}}}
\newcommand{\repourlfull}{https://github.com/Helmholtz-AI-Energy/frog}
\newcommand{\repourlshort}{https://github.com/Helmholtz-AI-Energy/frog}
\newcommand{\repourl}{\href{\repourlfull}{\repourlshort}}

\newcommand{\Ocal}{\mathcal{O}}
\newcommand{\ops}{\text{ops}}

\newcommand{\fgsum}[1][V]{g^+_#1}
\newcommand{\fgmean}[1][V]{\overline{g_#1}}
\newcommand{\frog}{P_{U}(\nabla f)}

\newcommand{\citet}[1]{\cite{#1}}

\title{Beyond Backpropagation: Optimization with Multi-Tangent Forward Gradients}

\author{
\IEEEauthorblockN{
    Katharina Flügel\IEEEauthorrefmark{1}\IEEEauthorrefmark{2}, 
    Daniel Coquelin\IEEEauthorrefmark{1}\IEEEauthorrefmark{2}, 
    Marie Weiel\IEEEauthorrefmark{1}\IEEEauthorrefmark{2}, 
    Charlotte Debus\IEEEauthorrefmark{1}, 
    Achim Streit\IEEEauthorrefmark{1}, 
    Markus Götz\IEEEauthorrefmark{1}\IEEEauthorrefmark{2}}
\IEEEauthorblockA{
    \IEEEauthorrefmark{1}Karlsruhe Institute of Technology (KIT), Scientific Computing Center (SCC)
    \IEEEauthorrefmark{2}Helmholtz AI}
}

\maketitle

\newcommand{\ieeecopyrightintro}{Accepted version of the final article published in:}
\newcommand{\ieeecopyrightnotice}{2025 International Joint Conference on Neural Networks (IJCNN) \textbar{} 979-8-3315-1042-8/25 ©2025 IEEE \textbar{} DOI: \href{https://doi.org/10.1109/IJCNN64981.2025.11227446}{10.1109/IJCNN64981.2025.11227446}}
\begin{tikzpicture}[remember picture,overlay]
\node[anchor=center, align=center, text width=20cm] at ([yshift=2cm] current page.south)
      {\footnotesize \ieeecopyrightintro{}\\[-1mm] \ieeecopyrightnotice};
\end{tikzpicture}%
\begin{abstract}
    The gradients used to train neural networks are typically computed using backpropagation.
While an efficient way to obtain exact gradients, backpropagation is computationally expensive, hinders parallelization, and is biologically implausible.
Forward gradients are an approach to approximate the gradients from directional derivatives along random tangents computed by forward-mode automatic differentiation.
So far, research has focused on using a single tangent per step.
This paper provides an in-depth analysis of multi-tangent forward gradients and introduces an improved approach to combining the forward gradients from multiple tangents based on orthogonal projections.
We demonstrate that increasing the number of tangents improves both approximation quality and optimization performance across various tasks.

\end{abstract}


\section{Introduction}
\label{sec:introduction}

Neural networks are typically trained by minimizing a loss using gradient descent, that is, updating the network parameters by stepping in the opposite direction of their gradient.
Backpropagation~\cite{rumelhartLearningRepresentationsBackpropagating1986,linnainmaa1970representation} is widely used to compute these gradients efficiently and accurately but comes with several drawbacks.
Taking about twice the time of the forward pass~\cite{kaplanScalingLawsNeural2020}, it accounts for a large fraction of the training time, directly contributing to the consumed energy~\cite{debusReportingElectricityConsumption2023}.
Its forward-backward dependencies furthermore lead to suboptimal memory access patterns~\cite{craftonLocalLearningRRAM2019} and hinder parallelization~\cite{belilovskyDecoupledGreedyLearning2020a}.
For example, when pipelining the training over the network layers, these dependencies can lead to pipeline bubbles and under-utilization~\cite{narayananPipeDreamGeneralizedPipeline2019,huangGPipeEfficientTraining2019}.
Besides these practical issues, backpropagation is incompatible with the biological motivation behind artificial neural networks, as biological neural networks lack such backward pathways to communicate update information~\cite{bengioBiologicallyPlausibleDeep2016}.

To overcome these issues, there has been much interest in alternative approaches to train neural networks.
One promising approach, the recently proposed \emph{forward gradient}~\cite{baydinGradientsBackpropagation2022,silver2022learning}, uses forward-mode automatic differentiation to compute a directional derivative along a random tangent, approximating the gradient in a single forward pass.
This method has shown great potential for training both fully-connected and convolutional neural networks.
However, with increasing dimension, i.e., model parameter count, the approximation quality decreases while the variance of the forward gradient increases~\cite{belouzeOptimizationBackpropagation2022,ren2023scaling}.
By sampling a random direction, the forward gradient acts as a Monte-Carlo approximation of the gradient.
We thus expect that increasing the number of tangents will improve the approximation quality, enabling forward gradients to scale to larger dimensions and improving their robustness.
Using multiple tangents has been previously mentioned~\cite{baydinGradientsBackpropagation2022,silver2022learning} but has not yet been analyzed in detail or evaluated experimentally.

In this work, we aim to fill this gap by taking a detailed look at how using multiple tangents affect forward gradients.
We find that additional tangents improve the forward gradient with a tradeoff between accuracy and compute.
Specifically, we address the following questions:
\begin{enumerate}[label=\textbf{RQ\arabic*}, leftmargin=*, ref=\textbf{RQ\arabic*},noitemsep,topsep=-10pt]
    \item \label{rq:multi_tangent} Do multiple tangents improve the forward gradient?
    \item \label{rq:combination} How should the forward gradient information from multiple tangents be combined?
    \item \label{rq:scalability} Can multi-tangent forward gradients scale to state-of-the-art architectures?
    \item \label{rq:tradeoff} What are the trade-offs of using multiple tangents?
\end{enumerate}
\section{Related Work}
\label{sec:related_work}

Forward gradients \cite{baydinGradientsBackpropagation2022,silver2022learning} are an approach to approximate the gradient with the directional derivative along a random direction, resulting in an unbiased estimator of the gradient.
A follow-up study \cite{belouzeOptimizationBackpropagation2022} further analyzes the forward gradient on the Beale and Rosenbrock functions, revealing significant shortcomings in high dimensions. 
It has been shown that the variance of the forward gradient estimation increases with the dimension, and activity perturbation reduces this variance compared to weight perturbation~\cite{ren2023scaling}.
Using a custom architecture based on MLPMixer~\cite{tolstikhinMLPMixerAllMLPArchitecture2021} and combining activity perturbation with numerous local losses allows scaling up to ImageNet~\cite{dengImageNetLargescaleHierarchical2009a}.
A framework~\cite{fournierCanForwardGradient2023} to combine different gradient targets and gradient guesses, i.e., tangents, demonstrates significant improvements from using local update signals as tangents instead of random noise, scaling up to ResNet18~\cite{heDeepResidualLearning2016} and ImageNet32~\cite{chrabaszczDownsampledVariantImageNet2017a}.

Besides forward gradients, numerous other approaches seek to train neural networks without backpropagation, often to overcome its biological implausibility.
Feedback Alignment~\cite{lillicrapRandomSynapticFeedback2016} aims to enhance biological plausibility and solve the so-called weight transport problem~\cite{grossbergCompetitiveLearningInteractive1987} by communicating feedback signals backward along random, untrained feedback paths.
Follow-up works~\cite{noklandDirectFeedbackAlignment2016,frenkelLearningFeedbackFixed2021,flugelFeedForwardOptimizationDelayed2023} replace this backward step with direct forward communication to each layer, effectively addressing the update locking problem~\cite{jaderbergDecoupledNeuralInterfaces2017}.
Forward gradients have been combined with direct feedback alignment~\cite{noklandDirectFeedbackAlignment2016} and momentum to reduce the gradient variance~\cite{bachoLowvarianceForwardGradients2024}.
The forward-forward algorithm~\cite{hintonForwardForwardAlgorithmPreliminary2022} trains neural networks using two forward passes---one on positive data and another on negative data---while applying local updates to differentiate between them.
Similarly, PEPITA~\cite{dellaferreraErrordrivenInputModulation2022a} applies a second forward pass on a randomly modulated input, which has been shown to be a special case of forward-forward~\cite{srinivasanForwardLearningTopFeedback2023}.
%
Synthetic gradients~\cite{jaderbergDecoupledNeuralInterfaces2017,czarneckiUnderstandingSyntheticGradients2017} approximate the gradient by training layer-local models that predict the gradient using only local information, effectively decoupling the layers.

\section{Multi-Tangent Forward Gradients}
\label{sec:main}

\subsection{Background}
\label{sub:background}

A typical neural network consists of \emph{layers} $l_1,\dots, l_L$, each processing an input $h_{i-1}$ from the previous layer, starting from the network input $h_0$, and passing the output $h_i$ on to the next layer.
Each layer depends on parameters $\theta_i$, which are adjusted during training.
A layer can thus be seen as a function $l_i: \Theta_i \times H_{i-1} \to H_i, (\theta_i, h_{i-1})\mapsto h_i$, mapping the parameter space $\Theta_i$ and the input space $H_{i-1}$ to the output space $H_i$.
In the \emph{forward pass}, an input $x=h_0$ is mapped to an output $y=h_L$ by successively applying these layers.
Using this layer definition, we can define the entire model as $m:\Theta\times X\to Y$ with $\Theta=\Theta_1\times\dots\times\Theta_L$, $X=H_0$, and $Y=H_L$, where 
$m(\theta, x)\mapsto l_L(\theta_L, l_{L-1}(\dots, l_1(\theta_1,x)))$ with $\theta=(\theta_1,\dots,\theta_L)$.

To train the model in a supervised fashion, we need an optimization metric, the \emph{loss function} 
$\mathcal{L}:Y\times Y\to\mathbb{R}$, which measures how much the model output $y=m(\theta, x)$ for an input $x$ differs from the corresponding target $y^*$.
We then train the model by minimizing the loss $\mathcal{L}(y, y^*)$, i.e., minimizing the objective function $f:\Theta\times X\times Y\to\mathbb{R}, (\theta, x, y^*)\mapsto \mathcal{L}(m(\theta, x), y^*)$ for a set of input samples $(x, y^*)$ by varying the parameters $\theta$.
A common method to minimize such a function is \emph{gradient descent}, where we take small steps in the opposite direction of the gradient 
\begin{align}
    \nabla f(\theta, x, y^*)=\frac{\partial f(\theta, x, y^*)}{\partial \theta}
\end{align}
of $f(\theta, x, y^*)$ with respect to the parameters $\theta$.
During training, the parameters are iteratively updated as
\begin{align}
    \label{eq:sgd_update_step}
    \theta\gets \theta -\eta\nabla f(\theta, x, y^*)
\end{align}
using a scalar learning rate $\eta$.
More advanced optimizers like Adam~\cite{kingmaAdamMethodStochastic2015} introduce additional terms, such as momentum and parameter-wise adaptive learning rates, while retaining the core idea of computing the gradient and stepping in the opposite direction.
Instead of using the exact gradient $\nabla f$, one can also train the model with an \emph{approximate gradient} $g$ like the forward gradient by replacing $\nabla f$ in \cref{eq:sgd_update_step} with $g$.

Computing the gradient $\nabla f$ is a non-trivial task.
Multiple methods exist, for example, by analytically deriving a closed-form solution or numerically approximating the gradient using the finite differences approach.
However, most are impractical for complex neural networks due to inefficient computation and numerical instability~\cite{baydinAutomaticDifferentiationMachine2018}.
This is why automatic differentiation, specifically backpropagation~\cite{rumelhartLearningRepresentationsBackpropagating1986,linnainmaa1970representation}, is commonly used to compute gradients for neural network training.

\emph{Automatic differentiation (AD)} is a computationally efficient approach for computing partial derivatives by repeatedly applying the chain rule and simplifying intermediate results to numerical values.
The objective function $f(\theta, x, y^*)$ is a function composition whose derivative can be calculated using the chain rule.
As an example, let us consider a function composition $f:\mathbb{R}\to\mathbb{R}, x\mapsto g_3(g_2(g_1(x)))$ with $g_i:\mathbb{R}\to\mathbb{R}$ and let $h_i=(g_i\circ\dots\circ g_1)(x)$.
The derivative of $y=f(x)$ with respect to $x$ is
\begin{align}
    \frac{\partial y}{\partial x} = \frac{\partial y}{\partial h_2} \frac{\partial h_2}{\partial h_1} \frac{\partial h_1}{\partial x}.
\end{align}
This product can be computed starting from either end, resulting in a forward- and a reverse-mode of AD. 
The forward-mode initializes $\partial x/\partial x=1$ and successively computes $\partial h_i/\partial x$ from $\partial h_{i-1}/\partial x$, while reverse-mode AD initializes $\partial y/\partial y=1$ and computes $\partial y/\partial h_i$ from $\partial y/\partial h_{i+1}$.
%
Higher dimensions are more complex as the derivative of a function $f:\mathbb{R}^n\to\mathbb{R}^m, x\mapsto (f_1(x), \dots, f_m(x))$ is the $m\times n$ Jacobi matrix $\mathbf{J}_f$.
To compute all entries of $\mathbf{J}_f$, forward-mode requires $n$ passes, as it needs to initialize $n$ different $\partial x/\partial x$ for all $n$ dimensions of $x$.
This corresponds to computing the Jacobi-vector product $\mathbf{J}_fv$ for $v\in\mathbb{R}^n$.
Correspondingly, reverse-mode requires $m$ passes with different $\partial y/\partial y$, each computing the vector-Jacobi product $v^\top\mathbf{J}_f$ for a $v\in\mathbb{R}^m$.

The time complexity of evaluating $f$ and computing $k$ passes in either forward- or reverse-mode is $\omega\cdot\text{ops}(f)$, where $\text{ops}(f)$ is the time required for evaluating $f$ (inference), with $\omega\leq1+1.5k$ for forward-mode and $\omega\leq1.5+2.5k$ for reverse-mode AD~\cite{griewankEvaluatingDerivativesPrinciples2008}.
As obtaining the full Jacobi matrix $\mathbf{J}_f$ requires $k=n$ forward-mode or $k=m$ reverse-mode passes, the decision between them depends on the dimensions $n$ and $m$.
When training a neural network with many parameters (large $n$) by minimizing a scalar loss ($m=1$), computing the exact gradient $\nabla f$ is much more efficient with reverse-mode AD than forward-mode AD.
The special case of reverse-mode AD for $m=1$ is also known as \emph{backpropagation (BP)}.

While backpropagation is more efficient than forward-mode AD for computing exact high-dimensional gradients $\nabla f$, it comes with several drawbacks, such as forward, update, and backward locking~\cite{jaderbergDecoupledNeuralInterfaces2017}, which limit both its biological plausibility and the potential for model parallelism and pipelining.
An alternative to computing exact gradients in reverse-mode are approximate gradients based on forward computation.

\subsection{Forward Gradient}
\label{sub:forward_gradient}
\begin{figure}[tb]
    \centering
    \includegraphics[scale=0.35]{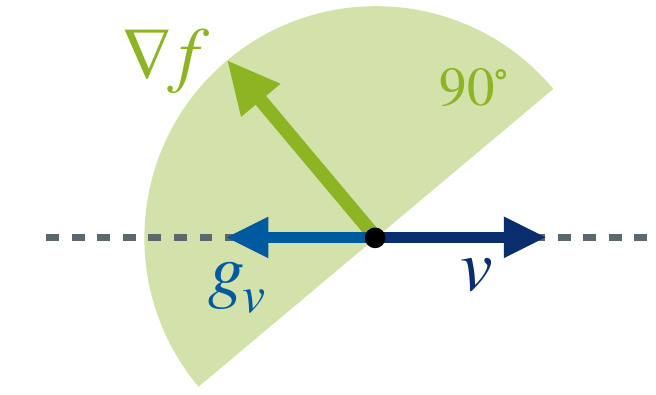}
    \caption{The forward gradient $g_v$ for the tangent $v$ is a projection of the gradient $\nabla f$ on the (1D) subspace spanned by $v$. It is by definition always within \ang{90} of $\nabla f$ and thus a descending direction of $f$.}
    \label{fig:forward_gradient}
\end{figure}

The \emph{forward gradient}~\cite{baydinGradientsBackpropagation2022,silver2022learning} is an approach to approximate the gradient $\nabla f$ by multiplying the directional derivative $\partial f/\partial v = \nabla f \cdot v$ along a tangent $v\in\mathbb{R}^n$ with $v$.
It can be computed efficiently without backpropagation using a single forward-mode AD pass to compute $\nabla f \cdot v$.
\begin{definition}[Forward Gradient]\label{definition:forward_gradient}
    The forward gradient $g_v$ for a tangent $v\in\mathbb{R}^n$ is defined as
    \begin{align}
        g_v = (\nabla f \cdot v)v.
        \label{eq:forward_gradient}
    \end{align}
\end{definition}
%
Forward gradients have multiple useful properties, also visualized in \Cref{fig:forward_gradient}.
%
\begin{property}[Unbiased Estimator]
    The forward gradient $g_v$ is an unbiased estimator of $\nabla f$, i.e., $\mathbb{E}[g_v]=\nabla f$, when sampling the tangent $v$ randomly with iid. entries $[v]_i$ with zero mean and unit variance~\cite{baydinGradientsBackpropagation2022,silver2022learning}.
\end{property}
%
\begin{property}[Within \ang{90}]\label{property:fg_90_deg}
    The forward gradient $g_v$ has non-negative dot-product with $\nabla f$ and is thus within \ang{90} of $\nabla f$ for any tangent $v$ if $g_v\neq 0$~\cite{silver2022learning}.
\end{property}
\begin{property}[Descending Direction]\label{property:desc_dir}
    The forward gradient is always a descending direction of $f$~\cite{silver2022learning}. 
\end{property}
\Cref{property:desc_dir} follows directly from \Cref{property:fg_90_deg}.
\begin{property}\label{property:stfg_span}
    The forward gradient $g_v:\mathbb{R}^n\to\text{span}(v)$ maps a concrete $\nabla f\in\mathbb{R}^n$ onto the one-dimensional subspace $\text{span}(v)$ spanned by $v\neq0$.
    If $v$ is a unit vector, $g_v$ is a projection.
\end{property}
\begin{proof}
    From \Cref{definition:forward_gradient}, it follows directly that $g_v\in\text{span}(v)$ since $\nabla f \cdot v$ is a scalar and $v\in\text{span}(v)$.
    If $v$ is a unit vector,
    \begin{align}
        (g_v \cdot v)v = ((\nabla f \cdot v)v \cdot v)v = (\nabla f \cdot v)v  = g_v,
    \end{align}
    and thus $g_v$ is a projection.
\end{proof}
Besides these beneficial properties, there are several drawbacks to consider.
While the forward gradient $g_v$ is an unbiased estimator, it suffers from a high variance that increases with the number of parameters $n$~\cite{belouzeOptimizationBackpropagation2022,ren2023scaling}, thus reducing the accuracy of the gradient approximation.
Specifically, the expected cosine similarity decreases with $\Ocal\left(1/n\right)$, approaching zero, i.e., orthogonality, for large $n$.
In contrast, the magnitude of $g_v$ scales with $\|v\|_2^2$, whose expectation increases with $\Ocal(n)$.
In summary, with increasing gradient dimension $n$, the forward gradient's direction becomes less accurate while its magnitude increases simultaneously, resulting in larger steps being taken in worse directions.
We confirm this observation experimentally in \Cref{sec:evaluation} and find that with increasing $n$, the single-tangent forward gradient struggles to optimize even trivial functions like the \emph{Sphere} function.

\subsection{Multi-Tangent Forward Gradients}
\label{sub:multi_tangent_fg}

When sampling tangents randomly, $g_v$ acts as a Monte-Carlo approximation of $\nabla f$.
Consequently, we expect the approximation quality to improve with more samples, leading us to the definition of a multi-tangent forward gradient (\labelcref{rq:multi_tangent}).
\begin{definition}[Multi-Tangent Forward Gradient]\label{definition:multi_tangent_fg}
    The multi-tangent forward gradient $g_V$ over a set of $k$ tangents $V=\{v_1, \dots, v_k\}$, $v_i\in\mathbb{R}^n$ is defined as 
    \begin{align}
        g_V=\bigoplus_{v_i\in V}g_{v_i}
        \label{eq:multi_tangent_fg}
    \end{align}
    for an aggregation operator $\oplus$.
\end{definition}
\begin{lemma}
    Let $V=\{v_1, \dots, v_k\}$ be $k$ linearly independent tangents $v_i\in\mathbb{R}^n$ and $U=\text{span}(V)\subseteq\mathbb{R}^n$ the subspace spanned by $V$.
    For any linear combination $\oplus$ applies $g_V\in U$.
\end{lemma}
\begin{proof}
    Per \Cref{property:stfg_span}, each single-tangent forward gradient $g_{v_i}$ lies in $\text{span}(\{v_i\})\subseteq U$.
    Thus, all $g_{v_i}$ are in $U$.
    Since vector spaces are closed under vector addition and scalar multiplication, $g_V\in U$ for any linear combination $\oplus$.
\end{proof}
A key question is how to combine the single-tangent forward gradients $g_{v_1}, \dots, g_{v_k}$ into a multi-tangent forward gradient, i.e., how to choose $\oplus$ (\labelcref{rq:combination}).
Two approaches have been suggested so far: 
\begin{definition}\label{definition:fg_sum}
    The multi-tangent forward gradient as a sum~\cite{baydinGradientsBackpropagation2022} is defined as
    \begin{align}
        \fgsum=\sum_{i=1}^k g_{v_i}= \mathbf{V}\mathbf{V}^\top\nabla f
        \label{eq:fg_sum}
    \end{align}
    with $\mathbf{V}=(v_1|\dots|v_k)$.
\end{definition}
\begin{definition}\label{definition:fg_mean}
    The multi-tangent forward gradient as average~\cite{silver2022learning} is defined as
    \begin{align}
        \fgmean=\frac{1}{k}\fgsum.
        \label{eq:fg_mean}
    \end{align}
\end{definition}
They correspond to choosing $\oplus=\sum$ and $\oplus=\overline{\openbox}$, respectively.
The resulting multi-tangent forward gradient differs only in its magnitude but not its direction, which is easily offset by the learning rate.

When using all $n$ standard basis vectors $e_1,\dots,e_n$ as tangents (i.e., $k=n$), $\fgsum$ corresponds to computing the exact gradient with full forward-mode AD, computing $\nabla f$ for one parameter at a time as described in \Cref{sub:background}.
However, when sampling the tangents randomly, e.g., $v_i\sim\mathcal{N}(0, I_n)$, $\fgsum$ does generally not equal $\nabla f$, even for $k=n$ tangents, as shown in \Cref{sub:eval:approximation_quality}.
This limitation arises from the aggregation method used to combine multiple forward gradients $g_{v_i}$: both average and sum are conical combinations, which restrict the result to the cone spanned by the $g_{v_i}$ and, consequently, the tangents $v_i$, as $g_{v_i}$ is always \mbox{(anti-)}parallel to $v_i$.
Moreover, they ignore the approximation quality of the different $g_{v_i}$ and their interrelations.
That is why even when using $k\geq n$ linearly independent tangents, $\fgsum$ and $\fgmean$ may fail to perfectly approximate $\nabla f$, despite having sufficient information to reconstruct the entire gradient.
    

\subsection{Improved Multi-Tangent Forward Gradients With Orthogonal Projection}
\label{sub:frog}

\begin{figure}[tb]
    \centering
    \includegraphics[width=.6\linewidth]{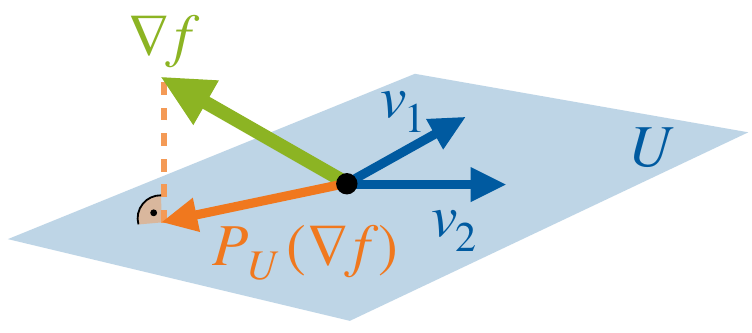}
    \caption{Orthogonal projection for $n=3$ and $k=2$. The tangents $v_1$ and $v_2$ span a two-dimensional plane $U$. The gradient $\nabla f$ does not lie within this plane, but its orthogonal projection $\frog$ provides the closest approximation of $\nabla f$ in $U$.}
    \label{fig:forward_gradient_projection}
\end{figure}

To improve the resulting multi-tangent forward gradient, we aim to use all information available from the single-tangent forward gradients.
\begin{definition}[Forward Orthogonal Gradient]\label{definition:frog}
    Let $V=\{v_1, \dots, v_k\}$ be $k$ linearly independent tangents with $U=\text{span}(V)$ and $\mathbf{V}=(v_1|\dots|v_k)$,
    then the \emph{orthogonal projection} of $\nabla f$ onto $U$ is defined as 
    \begin{align}
        \frog= \mathbf{V}(\mathbf{V}^\top \mathbf{V})^{-1}\mathbf{V}^\top\nabla f.
        \label{eq:frog}
    \end{align}
\end{definition}
\begin{lemma}
    $\frog$ is the most accurate approximation $g$ of $\nabla f$ in $U$ and minimizes $\|\nabla f - g\|_2$.
\end{lemma}\begin{proof}
    We adapt the proof of Proposition 9.6.6 in~\citet{lankhamLinearAlgebraIntroduction2016} and show that $\frog$ minimizes $\|\nabla f - g\|_2$ for every $g\in U$ using the Pythagorean Theorem and the monotonicity of $x^2$ and $\|\cdot\|_2\geq 0$.
    Let $g\in U$, then
    \begin{align}
        &\|\nabla f - \frog\|_2^2 \\
            \leq& \|\nabla f - \frog\|_2^2 + \|\frog - g\|_2^2\\
            \leq& \|\nabla f - \frog + \frog - g\|_2^2
            \leq \|\nabla f - g\|_2^2.
    \end{align}
    Thus, $\frog$ minimizes $\|\nabla f - g\|_2$ for $g\in U$, making it the most accurate approximation of $\nabla f$ in $U$.
\end{proof}
Compared to $\fgsum$ (\Cref{definition:fg_sum}), $\frog$ uses the same vector of directional derivatives $\mathbf{V}^\top\nabla f$ but ``corrects'' $V$ to an orthonormal basis using the inverse Gram matrix $G(V)^{-1}=(\mathbf{V}^\top \mathbf{V})^{-1}$ before aggregating it.
The orthogonal projection $\frog$ can thus be easily computed from the tangents' directional derivatives, providing a new multi-tangent forward gradient with improved accuracy (\labelcref{rq:combination}).
This is related to the geometrical interpretation given by~\citet{silver2022learning}; however, rather than requiring $V$ to be an orthonormal basis, our approach can handle arbitrary sets of linearly independent tangents.
\Cref{fig:forward_gradient_projection} illustrates $\frog$ with a three-dimensional example.

Approximating $\nabla f$ with $\frog$ solves two problems at once: the dependency of $\|g_v\|_2$ on $\|v\|_2$, discussed in \Cref{sub:forward_gradient}, and the suboptimal use of available information, outlined in \Cref{sub:multi_tangent_fg}.
As a result, $\frog$ yields the true gradient for $k=n$ linearly independent tangents and the best possible approximation for $k<n$ tangents.
Thus, $\frog$ is at least as accurate as the multi-tangent forward gradients $\fgsum$ and $\fgmean$ introduced previously.
Most importantly, $\frog=\nabla f$ when $\nabla f\in U$.


\subsection{Time Complexity of Forward Gradients}
\label{sub:complexity}

Computing a $k$-tangent forward gradient $\fgsum$ or $\fgmean$ has a time complexity of $\Ocal(k\cdot\ops(f))$ and aligns with the complexity of $k$ forward-mode AD passes (see \Cref{sub:background}).
Constructing $G(V)^{-1}$ for $\frog$ introduces an additional overhead of $\Ocal(k^2n)$, which is dominated by $\Ocal(k\cdot\ops(f))$ when $k$ is restricted to small constants, but must be considered for arbitrarily large $k$. 
In both cases, the time complexity scales with $k$, which limits the applicability of multi-tangent forward gradients since there is only about a factor of two to gain over backpropagation~\cite{kaplanScalingLawsNeural2020}, making any value of $k$ beyond small constants rather impractical (\labelcref{rq:tradeoff}).
Despite the computational overhead, it is crucial to study the potential gain from multi-tangent forward gradients as the theoretical overhead may be partially offset by more efficient implementations, beneficial memory access patterns, and additional parallelization opportunities.


\section{Evaluation}
\label{sec:evaluation}

We evaluate multi-tangent forward gradients in terms of their approximation quality and optimization performance.
We optimize multiple closed-form functions and train various neural networks on the image classification datasets MNIST~\cite{lecunMNIST1998} and CIFAR10~\cite{krizhevskyCIFAR2009} using cross-entropy loss and activity perturbation.
We split \num{10000} samples from the training set to be used for validation, both during the learning rate search and for early stopping during training.
We normalize the images but apply no other data augmentation.
To highlight the differences between the gradients, we use stochastic gradient descent without momentum or learning rate schedule.
All networks are trained for \num{200} epochs with early stopping after ten epochs without improvement, using a batch size of \num{64} for MNIST and \num{128} for CIFAR10.

We use activity perturbation when training neural networks with forward gradients, which has been found to reduce the variance of forward gradients since the activations are typically of lower dimension than the weights~\cite{ren2023scaling}.
Thus, $n$ corresponds to the total size of all activations.
That is, we compute the forward gradients with respect to the output activation of a layer and determine the gradients of the layer-internal weights analytically based on the activation gradient.
For this, we use the following definition of layers: for the fully-connected network, each fully-connected layer is considered together with its subsequent activation; for the residual network, each of the residual blocks is considered as a layer, with the processing both before and after the residual blocks grouped into another layer; finally, for the vision transformer, each encoder layer is considered as layer, with the input and output processing merged into the first and last layers, respectively.

To adjust for the different gradient norms, we perform a learning rate search for each variant using the asynchronous evolutionary optimizer Propulate~\cite{taubert2023massively} to search the logarithmic space $[\num{1e-6}, \num{1e0}]$.
All learning rates are provided together with the code.
%
All experiments were conducted using an Intel Xeon Platinum 8368 processor and an NVIDIA A100-40 GPU and implemented in Python 3.9.16 using PyTorch~\cite{paszkePyTorchImperativeStyle2019} 2.2.2 with CUDA 12.4.
Our code is publicly available at \repourl.

\subsection{Approximation Quality}
\label{sub:eval:approximation_quality}

First, we evaluate how well different forward gradients approximate the true gradient $\nabla f$ by considering their difference in terms of direction, via the cosine similarity, and magnitude, via the ratio $\|g_v\|_2/\|\nabla f\|_2$.
We use a fixed gradient $\nabla f=(1, \dots, 1)^\top\in\mathbb{R}^n$ and approximate it using the forward-gradient approaches introduced in \Cref{sec:main}:
the single-tangent forward gradient $g_v$ and multi-tangent forward gradients using either a sum ($\fgsum$), an average ($\fgmean$), or the orthogonal projection ($P_U$) for a set of $k$ tangents $V=\{v_1, \dots, v_k\}$, $v_i\sim\mathcal{N}(0, I_n)$.
Additionally, we test sum and average with normalized tangents $W=\{w_1, \dots, w_k\}$, $w_i=v_i/\|v_i\|^2$, labeled as $\fgsum[W]$ and $\fgmean[W]$, respectively.
Due to the rotational invariance of $\mathcal{N}(0, I_n)$, the choice of $\nabla f$ does not affect our results.

\begin{figure}[b]
    \centering
    \includegraphics[width=\linewidth]{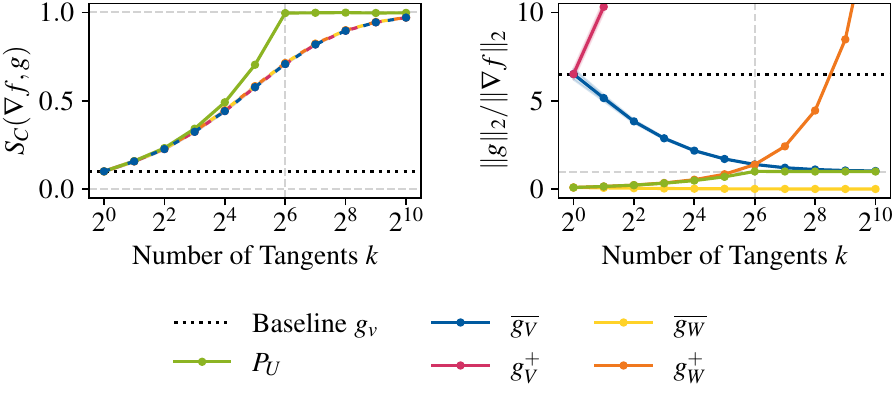}
    \caption{The approximation quality of different forward gradient approaches for $n=64$ in terms of cosine similarity and relative vector norm, mean over \num{1000} seeds. 
    As the cosine similarity of the conical combinations is identical, we use dashed lines to better visualize the overlapping curves.}
    \label{fig:eval_approximation_quality}
\end{figure}

\Cref{fig:eval_approximation_quality} illustrates our results for $n=64$ over \num{1000} seeds.
The dashed vertical line marks $k=64=n$; beyond this point, we could obtain a perfect approximation using full forward-mode AD.
The single-tangent baseline $g_v$ remains constant with $k=1$.
For the cosine similarity, we observe that all values lie between zero and one, confirming experimentally that the forward gradient is always within \ang{90} as shown in \Cref{property:fg_90_deg}.
As we increase the number of tangents, the cosine similarity increases, indicating a more accurate approximation of the true gradient's direction.
As expected, it is identical for all conical combinations $\fgsum$, $\fgmean$, $\fgsum[W]$, and $\fgmean[W]$ and approaches one only for $k\gg n$.
In contrast, the orthogonal projection $P_U$ is exact for $k\geq n$ and consistently outperforms the conical combinations, with the biggest gap around $k\approx n$.

The vector norm shows a notable difference between the conical combinations as the sum ($\fgsum$, $\fgsum[W]$) scales with $k$ while averaging ($\fgmean$, $\fgmean[W]$) over more tangents reduces the length.
Moreover, the length of each forward gradient $g_v$ scales with $\|v\|_2^2$.
This is evident in our results, as the length using the non-normalized tangent set $V$ scales with $\mathbb{E}[\|v\|_2^2]=n$ compared to the normalized set $W$.
For large $k$, $\fgmean$ approaches the correct length.
The orthogonal projection $P_U$ performs similarly to $\fgsum[W]$ for small $k$ but yields correct results for $k\geq n$ instead of significantly overestimating the gradient.

In summary, the orthogonal projection $P_U$ matches or outperforms all conical combinations in terms of both direction and magnitude, with the largest difference around $k\approx n$.
The conical combinations do not differ in their direction but only their length, which can be corrected by choosing an appropriate learning rate.
That is why we consider only the mean of forward gradients for non-normalized tangents $\fgmean$ in the remaining experiments.


\subsection{Optimizing Closed-Form Functions}
\label{sub:eval:math_optimization}

\begin{figure*}[tb]
    \centering
    \includegraphics[width=\linewidth]{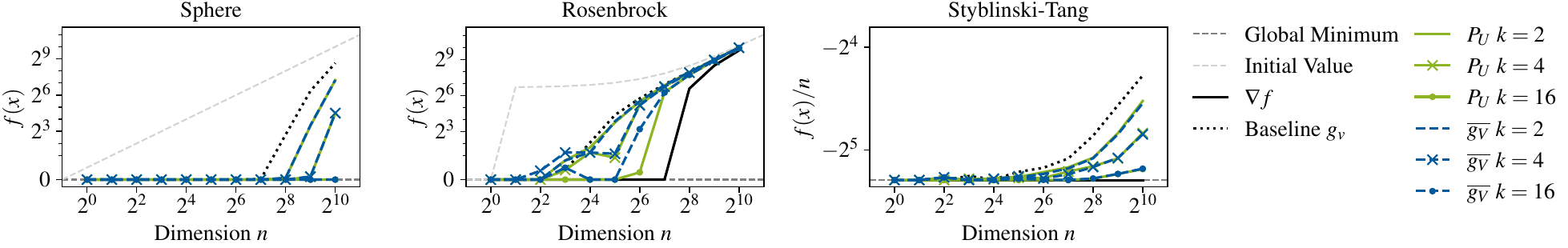}
    \caption{The best value found when minimizing $f:\mathbb{R}^n\to\mathbb{R}$ with different gradient approximations, mean over five random seeds. 
    To improve clarity for \emph{Styblinski-Tang}, we plot $f(x)/n$ instead, as the global minimum $-39.17n$ scales with $n$, and cut off the initial value (\num{0} for all $n$) to zoom in on the relevant data.}
    \label{fig:eval_math_optimization}
\end{figure*}

We first evaluate the practical application of multi-tangent forward gradients by minimizing three classical, multidimensional, differentiable closed-form functions with gradient descent. 
While the strictly convex \emph{Sphere} function is straightforward to optimize, the \emph{Rosenbrock}~\cite{rosenbrockAutomaticMethodFinding1960} and \emph{Styblinski-Tang} functions~\cite{styblinskiExperimentsNonconvexOptimization1990} pose harder problems, with a long, narrow, flat valley around the global minimum and many local minima, respectively.
We start at the initial values $x_\text{SP}^0=(-1, \dots, -1)^\top$, $x_\text{RB}^0=(-1, 0, \dots, 0)^\top$, and $x_\text{ST}^0=(0, \dots, 0)^\top$ correspondingly and perform up to \num{1000} update steps for \emph{Sphere} and \emph{Styblinski-Tang} and \num{25000} for \emph{Rosenbrock}, with early stopping after \num{50} steps without improvement.

We evaluate the best value reached using different gradient approximations for increasing dimensions $n$, giving the mean over five random seeds in \Cref{fig:eval_math_optimization}.
First, we observe that the single-tangent baseline $g_v$ struggles to optimize even trivial functions like the \emph{Sphere} for high-dimensional inputs.
Adding more tangents consistently improves the optimization result, effectively extending the range at which the forward gradient achieves performance comparable to the true gradient $\nabla f$.
The only exception is \emph{Rosenbrock}, where we find that for small $n$ and $k$, adding more tangents can actually decrease performance, especially for $\fgmean$.
Due to the complexity of optimizing \emph{Rosenbrock}, where a few missteps can quickly lead to divergence, and the learning rate increasing with $k$, $\fgmean$ can be more sensitive to the random seed.

Interestingly, the differences between $\fgmean$ and $P_U$ are less pronounced than those observed in \Cref{sub:eval:approximation_quality} when comparing the gradients themselves.
For \emph{Sphere} and \emph{Styblinski-Tang}, they seem to have no impact on the final result.
Only for \emph{Rosenbrock} do we observe a notable improvement of $P_U$ over $\fgmean$, where $P_U$ shows a more consistent improvement when adding more tangents.
This might be due to the fact that tangents sampled from $\mathcal{N}(0, I_n)$ are already very likely to be near orthogonal, limiting the potential impact of orthonormalization. 
On top of that, the difference in approximation quality observed in \Cref{sub:eval:approximation_quality} between $P_U$ and conical combinations is largest for $k$ close to $n$.
Yet the largest $k$ tested here is $k=16$, and for both \emph{Sphere} and \emph{Styblinski-Tang}, even the baseline $g_v$ converges to the minimum for $n\leq 16$.


\subsection{Non-Orthogonal Tangents}
\label{sub:eval:non_orthog_tangents}

\begin{figure}[tb]
    \centering
    \includegraphics[width=\linewidth]{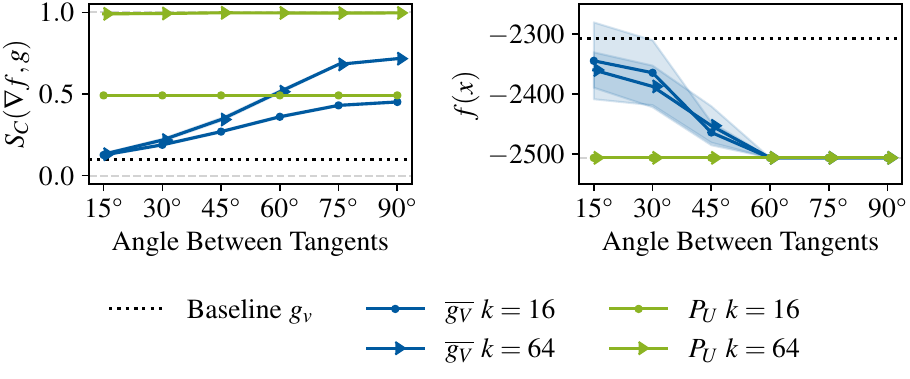}
    \caption{Cosine similarity (mean) and result on \emph{Styblinski-Tang} with $n=64$ (mean and ci) for different tangent angles $\alpha$ over \num{1000} and five seeds respectively.}
    \label{fig:eval_non_orthog_tangents}
\end{figure}

To investigate the benefits of orthogonal projection for arbitrary tangents, we study the behavior of $P_U$ and $\fgmean$ with non-orthogonal tangents.
For this, we construct a set of random tangents such that all tangents $v_2,\dots, v_k$ have a fixed angle $\alpha$ to the first tangent $v_1$.
The first tangent is again sampled from $\mathcal{N}(0, I_n)$. 
All further tangents $v_i$ are constructed by first sampling an $u_i\sim\mathcal{N}(0, I_n)$ and then rotating $u_i$ to $v_i$ in the plane spanned by $v_1$ and $u_i$ such that the angle between $v_i$ and $v_1$ is $\alpha$.
Intuitively, this samples the tangents randomly on an $\alpha$-cone around $v_1$.
As we decrease $\alpha$, the cone narrows, causing all tangents to be closer to $v_1$.

\Cref{fig:eval_non_orthog_tangents} gives the cosine similarity to $\nabla f$ and the best result on \emph{Styblinski-Tang} $n=64$ for different angles $\alpha$.
As expected, $P_U$ is entirely unaffected by $\alpha$ as it implicitly orthonormalizes the tangents and thus only depends on the dimensionality of their spanned subspace but not their angle.
In contrast, $\fgmean$ degrades as the cone around $v_1$ narrows until it is barely an improvement over using a single tangent.
This effect is more pronounced in the cosine similarity, where decreasing $\alpha$ immediately reduces $S_C(\nabla f, \fgmean)$. 
In contrast, the optimization results on \emph{Styblinski-Tang} continue to reach the optimum even down to \SI{60}{\degree}, with degradation only beginning for $\alpha\leq\SI{45}{\degree}$.
For $\alpha=\SI{90}{\degree}$, the result is comparable to that obtained when using tangents sampled from $\mathcal{N}(0, I_n)$ without rotation.
This is in line with the expectation that such tangents are near-orthogonal.


\subsection{Fully-Connected Neural Networks}
\label{sub:eval:fc_nn}

\begin{figure*}[tb]
    \centering
    \includegraphics[width=\linewidth]{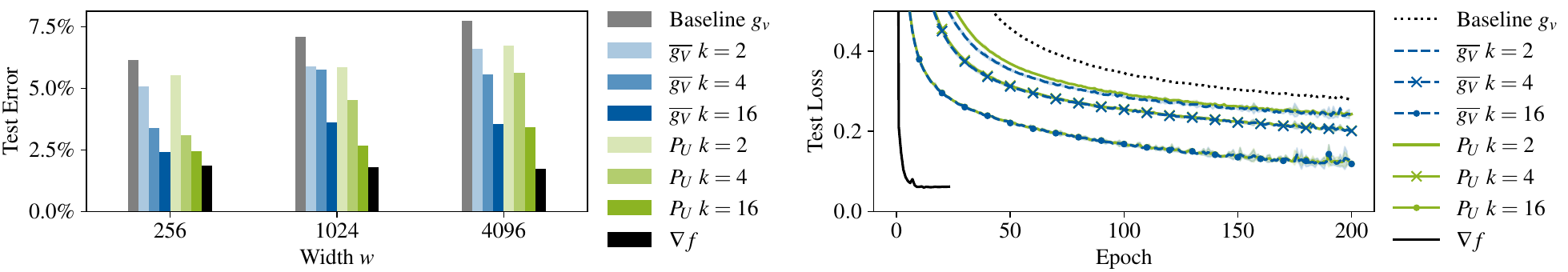}
    \caption{The best test error when training fully-connected networks with different hidden layer widths on MNIST and the test loss curve for width $w=4096$, where the gradient dimension is $n=2w+10$. Mean over three seeds.}
    \label{fig:eval_fc_nn}
\end{figure*}

We transfer our results to neural networks by training a fully-connected network with two hidden layers and ReLU activations on MNIST, flattening the image before passing it to the first layer.
We test three layer widths $w=\num{256}$, \num{1024}, and \num{4096} with different forward gradient variants.
Since we are using activity perturbation, the gradient dimension $n$ corresponds to the total size of all activations (including hidden), which correlates with the layer width $w$ as $n=2w+10$.
\Cref{fig:eval_fc_nn} gives the best test error after training for each layer width and the test loss curves for $w=4096$.
As in the previous experiments, we observe a clear improvement from using more tangents, with $k=16$ achieving results close to the true gradient $\nabla f$ despite $k\ll n$.
We again find only little practical difference between $P_U$ and $\fgmean$, although $P_U$ appears more robust and results in slightly smoother training.
Comparing different widths, we find that forward gradients can lose accuracy as the network size increases, whereas using $\nabla f$ yields marginally better accuracy with larger $w$.
This aligns with our previous observations that forward gradients with a fixed number of tangents become less accurate as we increase the input dimension $n$.
The loss curves clearly show that all tested forward gradient approaches require significantly more epochs than $\nabla f$, which converges within \num{25} epochs.
Using more tangents can both speed up the training process and improve the trained network's predictive performance. 
However, it still takes significantly more epochs than using the true gradient $\nabla f$ when both use appropriately tuned learning rates.


\subsection{Scaling to Larger Networks}
\label{sub:eval:sota_nn}

To analyze the scalability of multi-tangent forward gradients (\labelcref{rq:scalability}), we train a ResNet18~\cite{heDeepResidualLearning2016} and a small vision transformer (ViT)~\cite{dosovitskiyImageWorth16x162021} with six layers, four heads,  \num{256} hidden dimensions, and \num{512} MLP dimensions on MNIST and CIFAR10.
\Cref{table:eval_sota_nns} states the minimum test errors after training with different gradient approximations.
In line with our prior results, using more tangents reduces the test error, with $k=16$ achieving the best results among the tested forward gradients. 
Again, the distinction between the combination approaches is less clear and further complicated by the impact of the chosen learning rate since $\fgmean$ tends to be significantly larger than $P_U$, especially for small $k$.
While this can, in theory, be corrected for, determining a suitable learning rate for such complex tasks is non-trivial.
Suggesting the learning rate analytically to reduce the search space is thus an important avenue for further research.
Overall, a significant gap between the forward gradients and the true gradient remains, in particular for more complex tasks. 
Combining multi-tangent forward gradients with better tangent sampling approaches appears to be a promising approach to further reduce this gap.

\begin{table}[tb]
    \centering
    \caption{Best test error in \% after training with different (forward) gradients, mean over three seeds. Best value across forward gradients highlighted in bold.}
    \label{table:eval_sota_nns}
    \begin{tabular*}{\linewidth}{@{\extracolsep{\fill}}rrrrrr}
    \toprule
        
    && \multicolumn{2}{c}{\small\textbf{ResNet18}} & \multicolumn{2}{c}{\small\textbf{ViT}} \\
    &\multirow{2}{*}{\tikz{\node[below left, inner sep=1pt] (k) {$k$};%
    \node[above right,inner sep=1pt] (n) {$n$};%
    \draw (k.north west|-n.north west) -- (k.south east-|n.south east);}}& \textbf{MNIST} & \textbf{CIFAR10}& \textbf{MNIST} & \textbf{CIFAR10} \\
    & & $\num{11.2e3}$& $\num{11.8e3}$& $\num{64.0e3}$& $\num{83.2e3}$ \\
    \midrule
    $\nabla f$  &-- & \num{0.48}    & \num{25.79}   & \num{2.39}    & \num{40.29}\\
    \cmidrule(lr{0.7em}){1-6}
    $g_v$       &1  & \num{4.13}    & \num{63.63}   & \num{60.53}   & \num{71.10}\\
    \arrayrulecolor{lightgray}\cmidrule(lr{0.7em}){1-6}\arrayrulecolor{black}
    $\fgmean$   &2  & \num{4.32}    & \num{60.77}   & \num{50.73}   & \num{69.49}\\
    $\fgmean$   &4  & \num{2.98}    & \num{55.87}   & \num{44.57}   & \num{66.98}\\
    $\fgmean$   &16 & \num{1.86}    & \num{49.35}   & \numbf{36.06} & \num{63.68}\\
    \arrayrulecolor{lightgray}\cmidrule(lr{0.7em}){1-6}\arrayrulecolor{black}
    $P_U$       &2  & \num{3.33}    & \num{58.14}   & \num{49.11}   & \num{68.12}\\
    $P_U$       &4  & \num{3.14}    & \num{55.24}   & \num{44.08}   & \num{65.72}\\
    $P_U$       &16 & \numbf{1.84}  & \numbf{47.24} & \num{42.60}   & \numbf{61.07}\\
    \bottomrule
    \end{tabular*}
\end{table}


\section{Discussion}
\label{sec:discussion}

In this paper, we systematically investigate how using multiple tangents can improve forward gradients, addressing questions \labelcref{rq:multi_tangent} to \labelcref{rq:tradeoff}.

\paragraph*{Answer to \labelcref{rq:multi_tangent}: More tangents improve both gradient approximation and optimization.}
Our results confirm prior assumptions that aggregating a forward gradient over multiple tangents improves its performance.
Increasing the number of tangents $k$ consistently improves both the approximation of the gradient's direction (\Cref{sub:eval:approximation_quality}) and optimization performance (\Crefrange{sub:eval:math_optimization}{sub:eval:sota_nn}).
We demonstrate that the single-tangent forward gradient often struggles in high-dimensional spaces, even for simple functions like the \emph{Sphere}, consistent with prior observations \cite{belouzeOptimizationBackpropagation2022,ren2023scaling}.
By increasing $k$, we can alleviate this to a degree and scale to larger dimensions.


\paragraph*{Answer to \labelcref{rq:combination}: Orthogonal projection can yield more accurate forward gradients, especially for non-orthogonal tangents.}

Our study of different combination approaches shows that the orthogonal projection of the gradient onto the subspace spanned by the tangents can be computed with low overhead from the individual forward gradients and yields the most accurate gradient approximation (\Cref{sub:eval:approximation_quality}).
Somewhat surprisingly, this does not directly translate to better optimization performance; a simple average over the single-tangent forward gradients performs roughly on par with the orthogonal projection (\Crefrange{sub:eval:math_optimization}{sub:eval:sota_nn}).
We demonstrate that this is due to the tangent sampling method producing near-orthogonal tangents, which makes orthogonalization less impactful, and show that orthogonal projection significantly outperforms simpler combinations when one cannot rely on the tangents being near-orthogonal (\Cref{sub:eval:non_orthog_tangents}).


\paragraph*{Answer to \labelcref{rq:scalability}: Multi-tangent forward gradient can train ResNet18 and ViT.}

In \Cref{sub:eval:sota_nn}, we demonstrate that multi-tangent forward gradients can train residual networks and vision transformers, achieving significantly improved performance by using more tangents.
Nonetheless, a considerable gap to the true gradient $\nabla f$ remains.
Combining multi-tangent forward gradients with better tangent sampling holds great promise for further improving these results.


\paragraph*{Answer to \labelcref{rq:tradeoff}: Additional tangents are a trade-off between accuracy and compute.}

As discussed in \Cref{sub:complexity}, the time complexity of forward gradients scales linearly with the number of tangents.
From a theoretical perspective, we thus expect using any more than two tangents to result in a longer runtime than backpropagation.
In practice, the more advantageous memory-access patterns of forward passes may allow an optimized implementation to compete with backpropagation, even for small $k>2$.
Yet, scaling $k$ to $\Ocal(n)$ for large networks is unlikely to be feasible.
However, our findings show that it is not necessary to scale $k$ up to $n$; even a few additional tangents can significantly enhance results.
Optimizing the forward gradient implementation and measuring the wall-clock time with varying $k$ compared to backpropagation are important aspects to address in future work.


\paragraph*{Outlook and Future Work}

Multi-tangent forward gradients are largely independent of the gradient target and the tangents used, particularly when using orthogonal projection, as shown in \Cref{sub:eval:non_orthog_tangents}.
They can thus be easily integrated with approaches that replace the gradient target or tangents.
By utilizing multiple tangents, we could even combine multiple different tangent sampling approaches to leverage the advantages of each.
More advanced tangents based on, e.g., local losses, auxiliary networks, or adaptive sampling, may no longer be near orthogonal, thus benefiting further from orthogonal projection.
We are excited about the potential of combining multi-tangent forward gradients with ongoing research in the field.
By eliminating the backward pass, forward gradients enable more efficient parallelization of the training process.
The forward-only computation can help eliminate pipelining bubbles and increase pipeline utilization.
Beyond that, parallelizing over the tangents is a promising approach to mitigate computational overhead of multi-tangent forward gradients as the forward passes per tangent are independent and yield a single scalar value for the directional derivative.


\section*{Acknowledgment}
\addcontentsline{toc}{section}{Acknowledgment}
This work is supported by the Helmholtz Association Initiative and Networking Fund on the HAICORE@KIT partition.

\IEEEtriggeratref{21} 
\bibliography{references}
\bibliographystyle{IEEEtran}

\end{document}